\documentclass[12pt]{article}

\usepackage[T1]{fontenc}
\usepackage{mathptmx}

\usepackage{newtxtext,newtxmath}

\usepackage[utf8]{inputenc}
\usepackage{graphicx}
\usepackage{caption}
\usepackage{amsmath}
\usepackage[utf8]{inputenc}
\usepackage{hyperref}
\usepackage{graphicx}
\usepackage{multirow}
\usepackage{amsmath}
\usepackage[left=15mm,right=15mm,top=20mm,bottom=20mm]{geometry}
\usepackage[gen]{eurosym}
\usepackage{pdfpages}
\usepackage{amsmath}
\usepackage{graphicx}
\usepackage{float}
\usepackage{subcaption}
\usepackage{gensymb}
\usepackage{caption}
\usepackage{subcaption}
\usepackage{pdfpages}
\usepackage{nth}
\usepackage{subcaption}

\usepackage{sectsty}
\sectionfont{\large\bfseries}

\usepackage{ragged2e}
\usepackage[utf8]{inputenc}
\usepackage{hyperref}
\usepackage{graphicx}
\usepackage{multirow}
\usepackage{amsmath}
\usepackage[left=15mm,right=15mm,top=20mm,bottom=20mm]{geometry}
\usepackage[gen]{eurosym}
\usepackage{pdfpages}
\usepackage{amsmath}
\usepackage{graphicx}
\usepackage{float}
\usepackage{subcaption}
\usepackage{gensymb}
\usepackage{caption}
\usepackage{subcaption}
\usepackage{pdfpages}
\usepackage{nth}
\usepackage{listings}
\usepackage{color} 
\usepackage{abstract}

\definecolor{mygreen}{RGB}{28,172,0} 
\definecolor{mylilas}{RGB}{170,55,241}
\begin{document}

\begin{center}
\parindent=0in
\newcommand{\HRule}{\rule{\linewidth}{0.5mm}}



\vspace*{5mm}

\HRule \\[0.4cm]
{\Large{\textbf{From Problem to Solution: Bio-inspired 3D Printing for Bonding Soft and Rigid Materials via Underextrusions}}}\\[0.3cm]
\HRule \\[1cm]

\begin{minipage}{\textwidth}\large
\begin{center}
\begin{tabular}{c}
\begin{tabular}{c c}
\begin{minipage}[t]{3in}
\normalsize
\centering
\textbf{Arman Goshtasbi\footnotemark[1]}\\
SDU Soft Robotics \\
University of Southern Denmark\\
Odense, Denmark \\
argo@sdu.dk
\end{minipage}
&
\begin{minipage}[t]{3in}
\normalsize
\centering
\textbf{Luca Grignaffini\footnotemark[1]}\\
Department of Biomechanical Engineering\\
University of Twente\\
Enschede, The Netherlands\\
l.grignaffini@utwente.nl
\end{minipage}
\end{tabular} \\[1.5cm]

\begin{minipage}[t]{3in}
\normalsize
\centering
\textbf{Ali Sadeghi\footnotemark[2]}\\
Department of Biomechanical Engineering\\
University of Twente\\
Enschede, The Netherlands\\
a.sadeghi@utwente.nl
\end{minipage}
\end{tabular}
\end{center}
\end{minipage}

\footnotetext[1]{These authors contributed equally.}
\footnotetext[2]{Corresponding author.}

\end{center}

\vspace{5mm}

\begin{abstract}
\normalsize
\noindent Vertebrate animals benefit from a combination of rigidity for structural support and softness for adaptation. Similarly, integrating rigidity and softness can enhance the versatility of soft robotics. However, the challenges associated with creating durable bonding interfaces between soft and rigid materials have limited the development of hybrid robots. Existing solutions require specialized machinery, such as polyjet 3D printers, which are not commonly available. In response to these challenges, we have developed a 3D printing technique that can be used with almost all commercially available FDM printers. This technique leverages the common issue of underextrusion to create a strong bond between soft and rigid materials. Underextrusion generates a porous structure, similar to fibrous connective tissues, that provides a robust interface with the rigid part through layer fusion, while the porosity enables interlocking with the soft material. Our experiments demonstrated that this method outperforms conventional adhesives commonly used in soft robotics, achieving nearly 200\% of the bonding strength in both lap shear and peeling tests. Additionally, we investigated how different porosity levels affect bonding strength. We tested the technique under pressure scenarios critical to soft and hybrid robots and achieved three times more pressure than the current adhesion solution. Finally, we fabricated various hybrid robots using this technique to demonstrate the wide range of capabilities this approach and hybridity can bring to soft robotics.
\end{abstract}

\section{Introduction}

Animals seamlessly navigate and interact with their environment thanks to their physical intelligence and adaptive qualities found in natural tissues\cite{nature, KIM2013287}. This inherent capability has created a paradigm shift in robotics design through the emergence of soft robotics, which emulates the characteristics of biological organisms\cite{KIM2013287, rus2015design}. Over the past decade, numerous bio-inspired designs, including the intricacies of an octopus arm\cite{OctopusarmAli, Octopuscecilia}, the adhesion of a gecko's leg\cite{gecko}, the rhythmic motion of an earthworm\cite{das2023earthworm}, and the texture of snakeskin\cite{Snakeskin}, have emerged, expanding the possibilities for robotics tasks. 

Despite the recent emphasis on replicating soft tissues toward developing entirely soft robots, it is crucial to recognize rigidity's significant advantage in animal biology. For instance, vertebrates' endoskeletons provide support against gravity and mechanical loads and function as anchor points for soft tissues\cite{fleming2015building}. Beyond establishing a structural frame, rigidity plays a pivotal role in improving grasping capabilities via claws and nails\cite{hamrick2001development}, providing protective mechanisms, such as the skull and spine safeguarding essential organs and facilitating food grinding through teeth.  

Similar to biological structures, integrating rigidity into soft robotic designs can significantly enhance versatility and address the prevalent limitations in soft robotics\cite{culha2017design,buchner2023vision}. For instance, insufficient bending stiffness limits the soft robotics tasks involving handling heavy objects and horizontal manipulation- capabilities naturally supported by skeletal structures in animals \cite{Hybrid_stokes}. Additionally, the integration of rigid components can emulate the functional advantages seen in animal claws and nails, providing manipulation of both small and large objects\cite{nail_soft,crossiant}. Furthermore, the reliance of most soft robots on traditional, delicate electronic components, such as batteries, pumps, and motors, requires protective measures that current soft materials fail to provide\cite{Combustion,Hybrid_stokes}. Therefore, integrating rigid elements not only compensates for these shortcomings but also significantly broadens the functional scope of soft robotic systems.

While integrating rigid elements in soft robotics design offers a substantial advantage, fabricating such hybrid robots is challenging due to issues in bonding soft and rigid materials \cite{saldivar2023bioinspired}. The difference in material properties at the interface can lead to stress concentration and eventual debonding\cite{stress_conc_1,stress_conc_2,stress_conc_3}. To overcome this, soft robotics often uses adhesives like silicone glue to attach soft and rigid surfaces or connect tubes in pneumatic actuators\cite{zhang2020modular,ozioko2021sensact,tatari2020bending}. Although seemingly practical, this bonding method introduces design and repeatability limitations. Furthermore, the reliance on adhesive bonding remains vulnerable, as the adhesive layer often becomes the first point of failure in the robot's structure.

\begin{figure}[t]
    \centering
    \includegraphics[width=\textwidth]{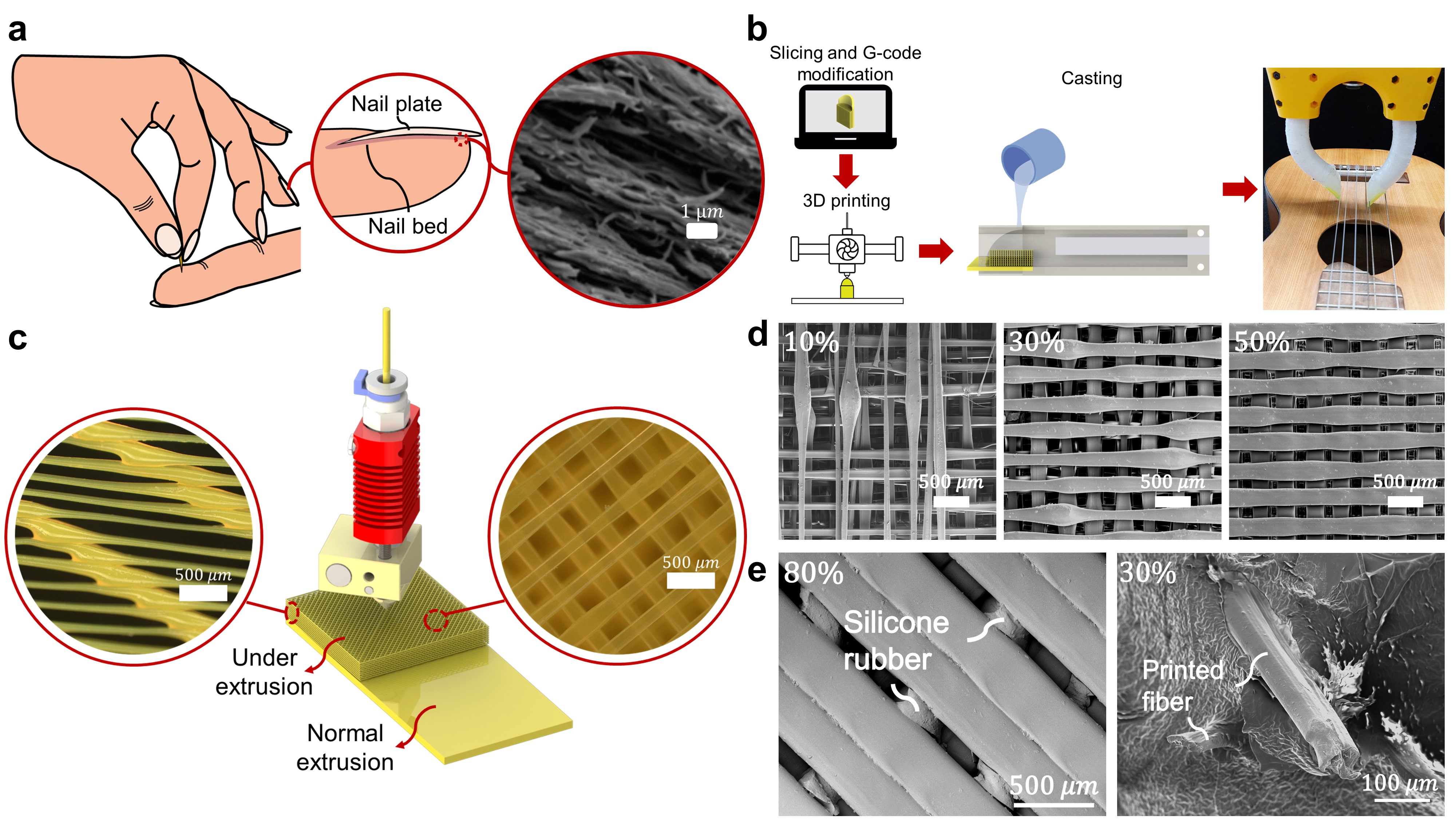}
    \caption{In nature, one of the many functions of connective tissue is to firmly bond rigid tissues to softer ones. In the human nail, the soft nail bed adheres to fibrous mesh keratine filaments that are present at the bottom of the rigid nail plate, which is shown in (\textbf{a}) (figure reproduced with permission from \cite{Nail_sem}). Example of the approach followed for the manufacturing of a hybrid gripper (\textbf{b}). 3D rendering of a sample used during lap shear bond tests with a zoomed-in microscopic view of the printed fibers (\textbf{c}). Scanning Electron Microscopy (SEM) images of samples printed at three different under-extrusion percentages (\textbf{d}). Conversely, in (\textbf{e}), two SEM images of samples printed at different under-extrusion percentages after silicone rubber was cast into the porous segments of the samples. It can be noticed how, at lower flow rates, the soft rubber completely envelops the printed fiber.}
    \label{fig:intro_fig}
\end{figure}

In animals' anatomy, however, rigid and soft tissues are intricately connected through robust yet flexible, fibrous connective tissues\cite{connective_tissue}. Thanks to their collagen fibers and porous structure, these tissues facilitate a resilient and flexible linkage within the body's structure. This is achieved through the interweaving of fibers within the tissues\cite{interwoven} that creates an interlocking mechanism to reduce the stress concentration by creating a stiffness gradient. Similarly, such bonding plays a significant role in the durable connection between soft and rigid structures in human anatomy. For instance, tendons are essential for connecting bones to muscles\cite{kamrani2019anatomy}, Sharpey's fibers, a matrix of connective tissues, are vital in anchoring the periosteum, the outer layer of bones, to the skin tissues\cite{aaron2012periosteal}, while the rigid fibrous keratin structure which is the rigid part of nails firmly interwove to the underlying skin layer, also known as the nail bed, for strong bonding, as shown in Fig.\ref{fig:intro_fig}\textbf{a}.

So far, increasing the surface of contact has been a technical solution for decreasing the stress concentration using advanced fabricating techniques, mainly additive manufacturing, to achieve more durable bonds between soft and rigid materials\cite{Combustion}e. For example, Al-Ketan et al.\cite{al2017mechanical} and Saldivar et al.\cite{saldivar2023bioinspired} utilized polyjet multi-material additive manufacturing to embed rigidity inside a matrix of soft materials and create stiffness gradient from soft to rigid for a better interface. Ma et al.\cite{ma2015hybrid} introduced a novel fabrication method combining fused deposition manufacturing (FDM) with material injection. This approach involved 3D printing anchors and hooks directly into a mold, effectively interlocked with the subsequently injected resin, ensuring a secure bond. Additionally, Rossin et al.\cite{rossing2020bonding} suggested another FDM technique to create grids as an interlocking mechanism for soft and rigid materials interface. 

While these studies present innovative approaches, they have limitations in soft robotics applications. For instance, polyjet multi-material additive manufacturing relies on expensive printers with a limited range of materials. Furthermore, the FDM techniques encounter challenges with the lower density of interlocking points compared to natural fibrous structures, which may affect the durability and effectiveness of the bond in soft robotics applications. 

Therefore, this study introduces a novel FDM 3D printing technique that utilizes the commonly encountered challenge of oozing to our advantage (Fig \ref{fig:intro_fig}). By employing this phenomenon, we have developed a method to replicate the complex, fibrous connective tissue structures in nature, creating a more resilient and durable connection between rigid and soft materials. This approach is particularly suited for the commonly used materials in soft robotics. We extensively compare the performance of our technique with traditional adhesive methods, demonstrating superior results in both shear and peeling tests. Additionally, due to the popularity of fluidic actuation and sensing in soft robotic systems, we also examined the airtightness of our bonding method via ballooning test. Our findings indicate that this new method not only overcomes the limitations of silicone glue but also significantly enhances the structural integrity and functionality of soft robotic systems.

\section{Results}

\subsection{Microscope Test}

The microscopy imaging revealed that the fiber diameter of the porous samples manufactured with the proposed method almost coincided with those predicted by theory \ref{eq1}-\ref{eq5}. As presented in Table \ref{tab:fiber}, the mean absolute error before predicted and experimental values of the diameter of the fibers is approximately 7 µm, with the sample printed at 100\% showing the largest deviation of 12 µm.

\begin{table}[ht]
\centering
\renewcommand{\arraystretch}{1.2}
\caption{\label{tab:fiber} Results of the microscopy measurements for samples printed at different flow rate percentages. The measured diameter of the PLA fibers is compared with the value predicted in equation (\ref{eq5}) in the methods section.}
\begin{tabular}{|c|c|c|c|}
\hline
Flow Rate Percentage (\%) & Predicted Diameter (µm) & Measured Diameter (µm) & Absolute Error (µm)\\
\hline
10 & 82 & 73 & 9\\
\hline
20 & 122 & 123 & 1\\
\hline
30 & 162 & 164 & 2\\
\hline
40 & 202 & 193 & 9\\
\hline
50 & 242 & 252 & 10\\
\hline
60 & 282 & 277 & 5\\
\hline
80 & 362 & 366 & 4\\
\hline
100 & 442 & 430 & 12\\
\hline
\end{tabular}

\end{table}

\subsection{Bonding Tests}

\begin{figure}[ht]
    \centering
    \includegraphics[width=\linewidth]{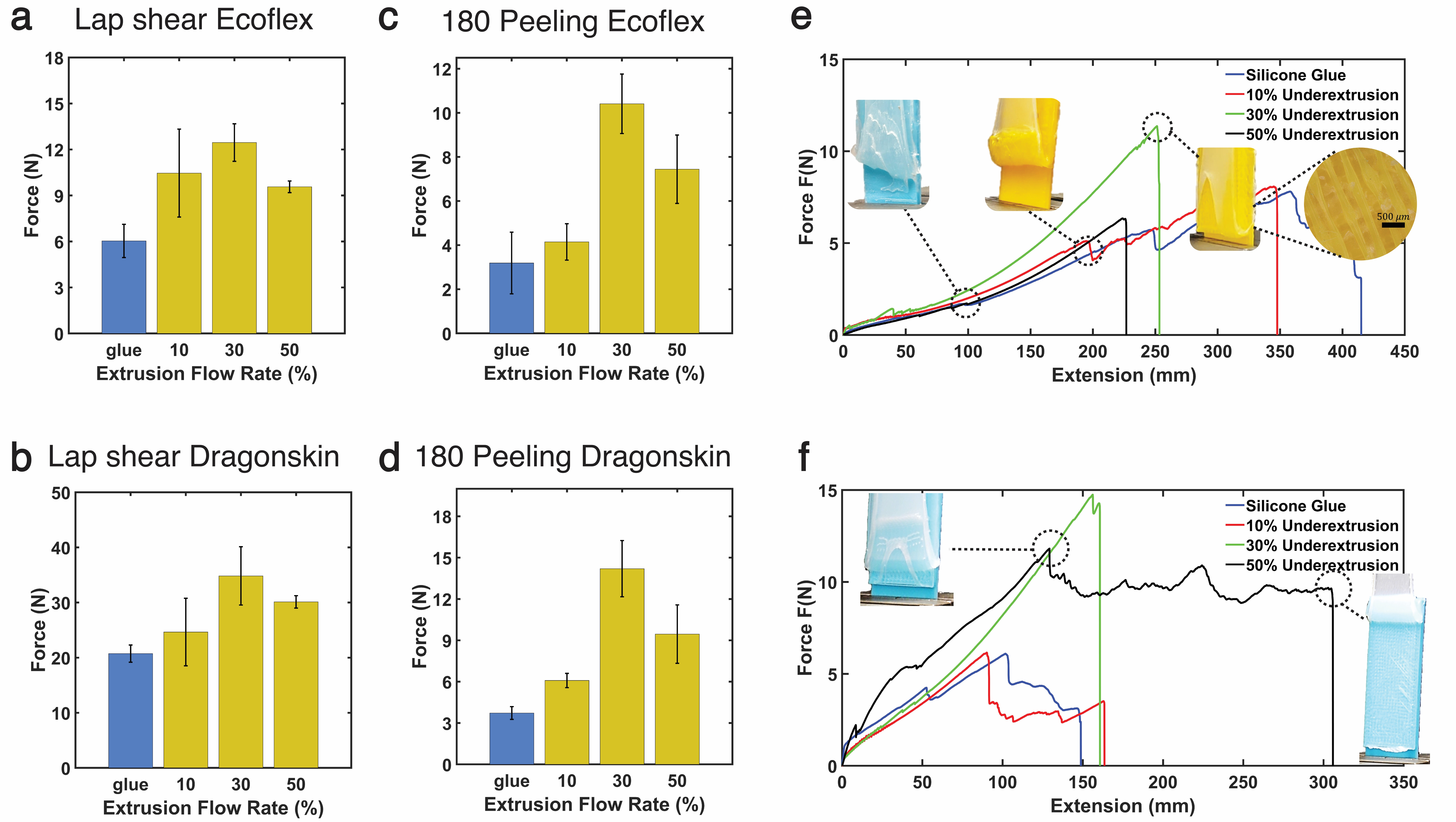}
    \caption{Results of the lap shear and peel-off tests. Barplot showing differences in leaking/breaking forces of hybrid PLA-Ecoflex 00-10 samples bonded with Sil-poxy adhesive (in blue) and 10, 30, and 50 underextrusion percentages (in yellow) for the lap shear (\textbf{a}), and peel-off bond tests (\textbf{c}). The same two tests were conducted for hybrid samples that bonded PLA to Dragon-Skin 10. The barplot illustrated in (\textbf{b}) refers to the lap shear test results, while barplot (\textbf{d}) refers to the peeling test. The values reported in the peel-off test bar plots referred to the initiation of breakage of the hybrid samples at the interface section of the free silicone and the bonding junction. In reality, the soft ecoflex 00-10 strip detached entirely from the samples in the cases of the silicone glue and 10\% samples, thus requiring a higher extension and overall force before the final rupture, as shown in (\textbf{e}). The same rupture mode happens when using Dragonskin 10 for the samples with a bonding section made at 50\% underextrusion, as seen in (\textbf{f}). 
    }
    \label{fig:result_bond}
\end{figure}

Both the lap shear and the peel-off test proved that our proposed technique offers a better bonding solution when manufacturing hybrid structures compared to commercially available adhesives such as silicone glue. As illustrated in Fig\ref{fig:result_bond}, all the different specimens made with our under-extrusion technique recorded higher values of the debonding force than the used silicone adhesive in both lap shear and 180 peeling test for both Ecoflex 00-10 and DragonSkin 10. When bonded to Ecoflex 00-10, the samples made at 30\% showed maximum values of 12.45 ± 1.22 N and 10.41 ± 1.35 N in the lap shear and 180 peeling test tests, respectively. The samples made at 10\% and 50\% respectively recorded 10.46 ± 2.87 N and 9.57 ± 0.38 N in the lap shear test, and 4.14 ± 0.82 N and 7.44 ± 1.55 N in the peeling test. The samples attached with the glue all showed an early onset of debonding, as they recorded much lower values of debonding forces at 6.03 ± 1.08 N in the lap shear test and 3.18 ± 1.39 N in the peel-off test. When using DragonSkin 10, the 30\% samples offered the best bonding solution, as they recorded values of 34.82 ± 5.29 N and 14.20 ± 2.03 N in the lap shear and peeling test, respectively. The 10\% samples recorded debonding forces of 24.64 ± 6.13 N for the lap shear test and 6.09 ± 0.51 N for the peeling test, while the 50\% samples registered 30.11 ± 1.12 N for the lap shear test and 9.45 ± 2.12 N. On the other hand, the samples bonded with Sil-Poxy silicone required much lower forces to detach from the PLA substrate (i.e., 20.74 ± 1.56 N for the lap shear test and 3.72 ± 0.46 N for the peeling test). 

\subsection{Pressure Test}

\begin{figure}[h!]
    \centering
    \includegraphics[width=\linewidth]{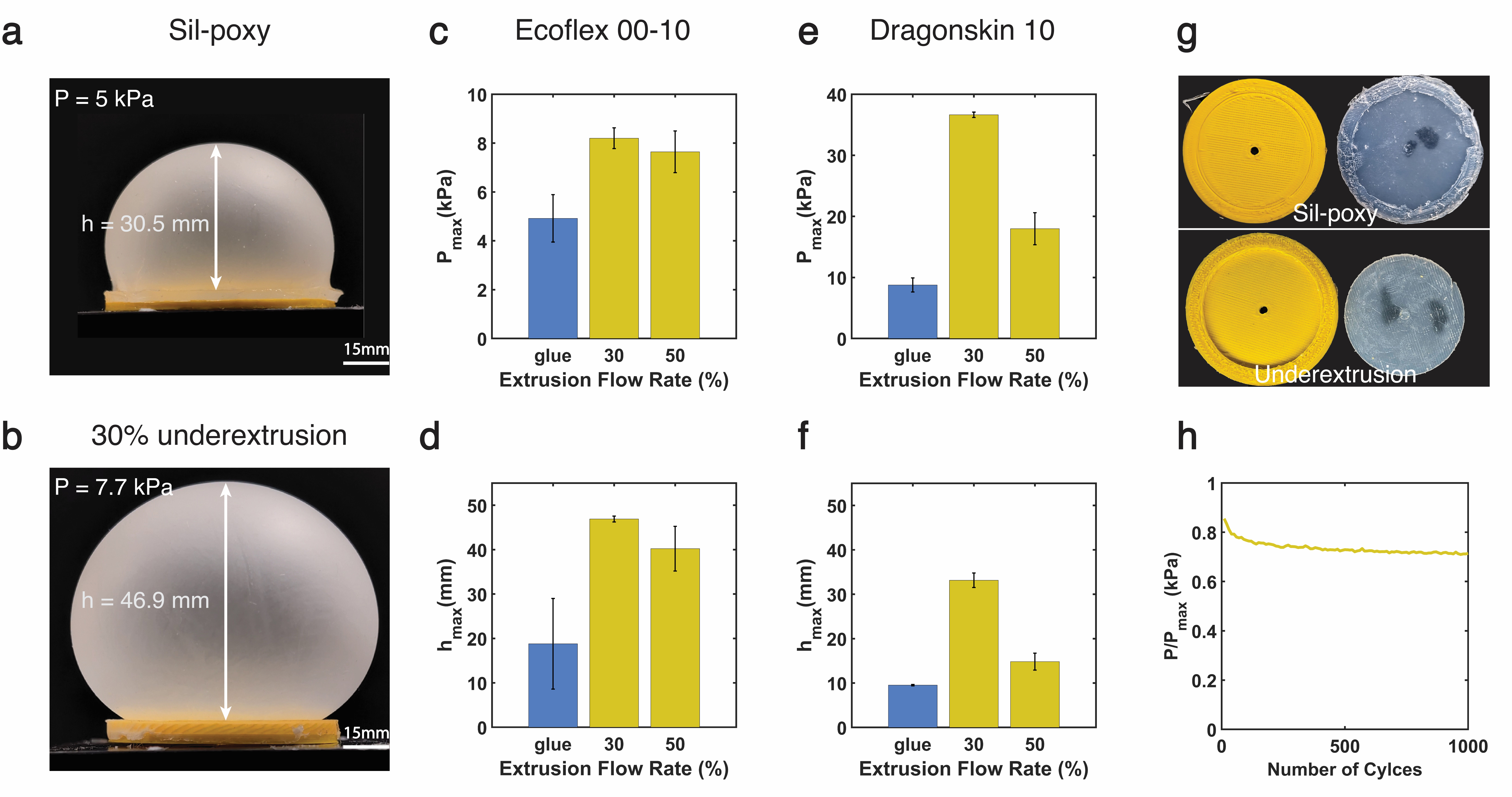}
    \caption{Results of the ballooning pressure test. Hybrid inflatable samples bonded with glue and underextrusion were pressurized until leaking/ruptured. (\textbf{a}), The Ecoflex 00-10 with Sil-poxy is captured before failure and leakage. (\textbf{b}) The Ecoflex bonding with underextrusion is shown. In this sample, the rubber layer undergoes plastic deformation without any rupture or leakage in the structure while standing at a higher pressure than glue (7.7 kPa). (\textbf{c}) The maximum recorded pressure of the Ecoflex 00-10 sample with three bonding methods. (\textbf{d}) The maximum deflection of Ecoflex 00-10 samples recorded for three scenarios. (\textbf{e}) The maximum recorded pressure of the Dragonskin10 sample with three bonding methods. (\textbf{f}) The maximum deflection of Dragonskin10 samples for three scenarios. (\textbf{g}) Sil-Poxy and underextrusion samples are shown after rupture. In the case of silicone-based glue, leakage is induced by the failure of the adhesive bond, which detaches the entire soft layer. When bonded with our method, leakage is induced by stress concentrations at the interface between the silicone and the porosity, thus leading to the rubber's failure and not the bonding. (\textbf{h}) The Ecoflex 00-10 cyclic test bonded to PLA with 30\% underextrusion for 1000 cycles and 80\% of the maximum value of highest pressure recorded at (\textbf{c}). 
    }
    \label{pressure}
\end{figure}

As detailed in the Experimental Setup and Materials section, given the prominence of pneumatics in soft robotics, we focused on the effect of underextrusion on the bonding strength between PLA and a layer of silicone rubber during pressure-induced expansion. As presented in Fig.\ref{pressure}, the findings indicate a significant increase in the pressure threshold and expansion peak for samples subjected to underextrusion compared to samples joined by adhesive for both Ecoflex 00-10 and Dragonskin 10.

During the Ecoflex 00-10 experiments, the pressure and expansion values of the underextruded samples (both 30\% and 50\%) exceeded those of the Sil-poxy bonded samples. As shown in Fig.\ref{pressure}c, pressures of 8.2 ± 0.4 kPa and 7.6 ± 0.8 kPa were achieved for 30\% and 50\% underextrusion, respectively, compared to 5 kPa for Sil-poxy. Additionally, we observed peak expansion points of 46.90 ± 0.66mm and 40.22 ± 6.84mm, respectively, leading to plastic deformation of the elastomer while maintaining bond integrity.

In the bonding test with Dragonskin 10, the 50\% underextrusion sample reached a pressure value of 18.0 ± 2.6 kPa, while the 30\% underextrusion sample reached 36.6 ± 0.4 kPa. The adhesive-bonded sample, however, withstood pressures below 8 kPa, indicating significantly lower pressure tolerance. Moreover, the peak expansion observed for the 30\% and 50\% underextruded samples were 33.14 ± 1.63 mm and 14.82 ± 1.90mm, respectively, surpassing the adhesive sample, which was limited to 9.54 ± 1.56mm (Fig.\ref{pressure} e-f).

Finally, as shown in Fig.\ref{pressure}h, the Ecoflex 00-10 samples bonded via 30\% underextrusion tolerated 80\% of the maximum pressure for 1000 cycles, with a slight pressure drop from 6.7 kPa to 5.7 kPa due to the plastic deformation of the silicone membrane.

\section{Discussion}
\subsection{Validation of the Technique}

\subsubsection{Microscopy}

Microscopic imaging showed that the widths of the extruded filaments made with the proposed technique follow the ones predicted by equation 5, as the computed average error is approximately 8.8 µm. 
The collagen fibers that comprise the connective tissue of animals in nature are characterized by an average diameter ranging between 1 µm and 10 µm \cite{DOUPLIK201347}. Using our method, we obtained fibers with a minimum average diameter of 73 µm by deliberately underextruding at a 10\% flow rate. Although reproducing the fibers of connective tissue found in nature was not our research's main goal, we believe this could be possible with our technique by further tuning various printing parameters such as print speed and print temperature. 
Additionally, we can see that by reducing the flow rate percentage, there is an increasing number of irregularities in different sections of the extruded filament, which is characterized by the alternation of thin sections with more pronounced bulges, as shown in the SEM images in Fig. \ref{fig:intro_fig}d, and in the optical microscopy images shown in Fig S1 of the supplementary material. This phenomenon could be caused by a physical phenomenon typically occurring in viscoelastic fluids, known as the Plateau–Rayleigh instability \cite{Plateu-Rayleigh}. By reducing the flow rate percentage, the surface tension forces prevail over the viscous forces, forming more significant instabilities in the extruded fibers. Another cause for the generation of these instabilities is strictly machine-related. Decreasing the flow rate reduced the amount of material by a factor $\gamma$, as shown in equation(\ref{eq4}). However, decreasing the flow rate does not affect the print speed, also known as the feed rate, which was kept at 80 mm/s for printing all the microscopy samples. The relatively high feed rate compared to the low flow rates may have caused stretching and thinning of the print filament, resulting in more extensive and more pronounced instabilities. 

All the test samples showed proper penetration of silicone rubber inside of the porosity, as proven by the illustrated in the SEM images shown in Fig.\ref{fig:intro_fig}e-f (Supplementary Fig S2), and in the optical microscopy snapshots in Fig S1 in the supplementary material. By decreasing the printing flow rate, which translates into a higher grade of porosity, a more significant amount of rubber can penetrate the porous segment. For instance, the silicone can completely envelop and surround the fibers printed at 30\%, as shown in Fig.\ref{fig:intro_fig}f. In contrast, the porous sample printed at 80\% flow rate shows visible gaps between the portions of the silicone rubber, as shown in Fig.\ref{fig:intro_fig}e. 

\subsubsection{Bond Tests}

After performing the lap shear and peel-off tests, we demonstrated that our proposed method offers a novel bonding solution between rigid materials and silicone rubber, significantly outperforming commercially available alternatives such as silicone rubber glue. Overall, the use of under-extrusion as a bonding technique shows to prevail over the silicone adhesive in both tests, with the 30\% flow rate showing the most considerable improvement (i.e., 106.2\% and 226.5\% in the lap shear and peeling tests for Ecoflex 00-10, and 68\% and 281\% for Dragon Skin 10, respectively) from the latter. 
When analyzing the modality of breakage of the hybrid samples made with PLA and Ecoflex 00-10, the entire silicone rubber strip was completely ripped off in all the specimens bonded with Sil-Poxy™ and with an under-extrusion flow rate percentage of 10\% in both tests. On the other hand, all the different samples characterized by higher flow rate percentages ruptured due to the failure of the silicone itself in correspondence with the interface section between the rubber and the porous segment, as can be seen in Fig.\ref{fig:result_bond}e. By increasing the percentage of under-extrusion, which translates into a lower degree of porosity, the impeded penetration of the soft material into the rigid segment causes stress concentrations that are responsible for the early onset of failure points at the material interfaces, ultimately reducing the recorded value of the force at the break. It is essential to clarify that the values of the reported forces in Fig.\ref{fig:result_bond} relate to the instances where the samples that were printed with an underextrusion percentage more larger than 10\% broke, as it equaled the onset of failure of all the samples. This choice was made as in most soft robotics applications (fluidic based), the first failures of the system occur with the onset leakage points, which may cause the robotic system as a whole to fail its purpose. We believe that the initial failures of the peeling tests in the proximity of the bonding interface section between the two materials can be classified as a potentially detrimental leaking point. 
In reality, the glued and 10\% samples recorded higher forces at their complete rupture, as shown in Fig\ref{fig:result_bond}e-f, which is logical considering the higher extension required to pull the entire silicone band. The difference in the observed rupture modes can be explained by the fact that by increasing the flow rate percentage and, therefore, the diameter of the extruded filament, the strength of the inter-layer PLA-PLA bond between the two different sections consequently increases. Thus, the soft material is not strong enough to detach the under-extruded segment from its underlying solid substrate when the flow rate is greater than 10\%, leading to its failure at the interface section (Supplementary Video S2).

\subsubsection{Pressure Test}

The Underextrusion 30\% bonding outperformed the Sil-poxy bonding for both Ecoflex 00-10 and Dragonskin 10, showing similar results to previous bonding tests. As presented in Fig.\ref{pressure}, for Ecoflex, we achieved 164\% more pressure and 249\% more maximum deformation. Moreover, we observed that both Underextrusion 30\% and Underextrusion 50\% reached plastic deformation. Underextrusion 30\% withstood the maximum possible air volume of the syringe system without any rupture, while Sil-poxy failed before reaching the plastic point (Video S3 in the Supplementary). The cyclic results in Fig.\ref{pressure}h indicate that despite the transient phase of the recorded pressure, the Underextrusion 30\% bonding is durable and resilient under cyclic loads. The pressure drop in the test is due to slight plastic deformation of the Ecoflex membrane.

For the Dragonskin membrane, we achieved a more significant pressure difference (4.5 times) and maximum deformation (3.5 times) between Underextrusion 30\% and Sil-poxy bonding. One main reason for this considerable pressure difference is the limited pressure tolerance of Sil-poxy in both Ecoflex and Dragonskin. In contrast, Underextrusion bonding allows Ecoflex to reach plastic deformation at pressures close to Sil-poxy's failure point, thus minimizing the pressure difference between the samples.

Finally, the reason that 30\% outperforms 50\% is similar to the bonding tests: the high density of fibers in the 50\% porosity prevents the silicone from penetrating properly inside the structure. This result is more significant when comparing the Ecoflex 00-10 and Dragonskin 10 samples, where the differences between 30\% and 50\% are smaller for Ecoflex 00-10 due to its lower viscosity, allowing easier penetration. Furthermore, in underextrusion cases, ballooning failures occur at the connection points between the non-bonded silicone section and the 3D-printed part, similar to the bonding tests. This failure is due to stress concentration from the rigid underextrusion to the soft silicone rubber. This is another reason why 30\% underextrusion performs better than 50\%, as the transition is smoother due to the lower stiffness of 30\% underextrusion.

\subsection{Hybrid Grippers}
\begin{figure}[ht]
\centering
\includegraphics[width=\linewidth]{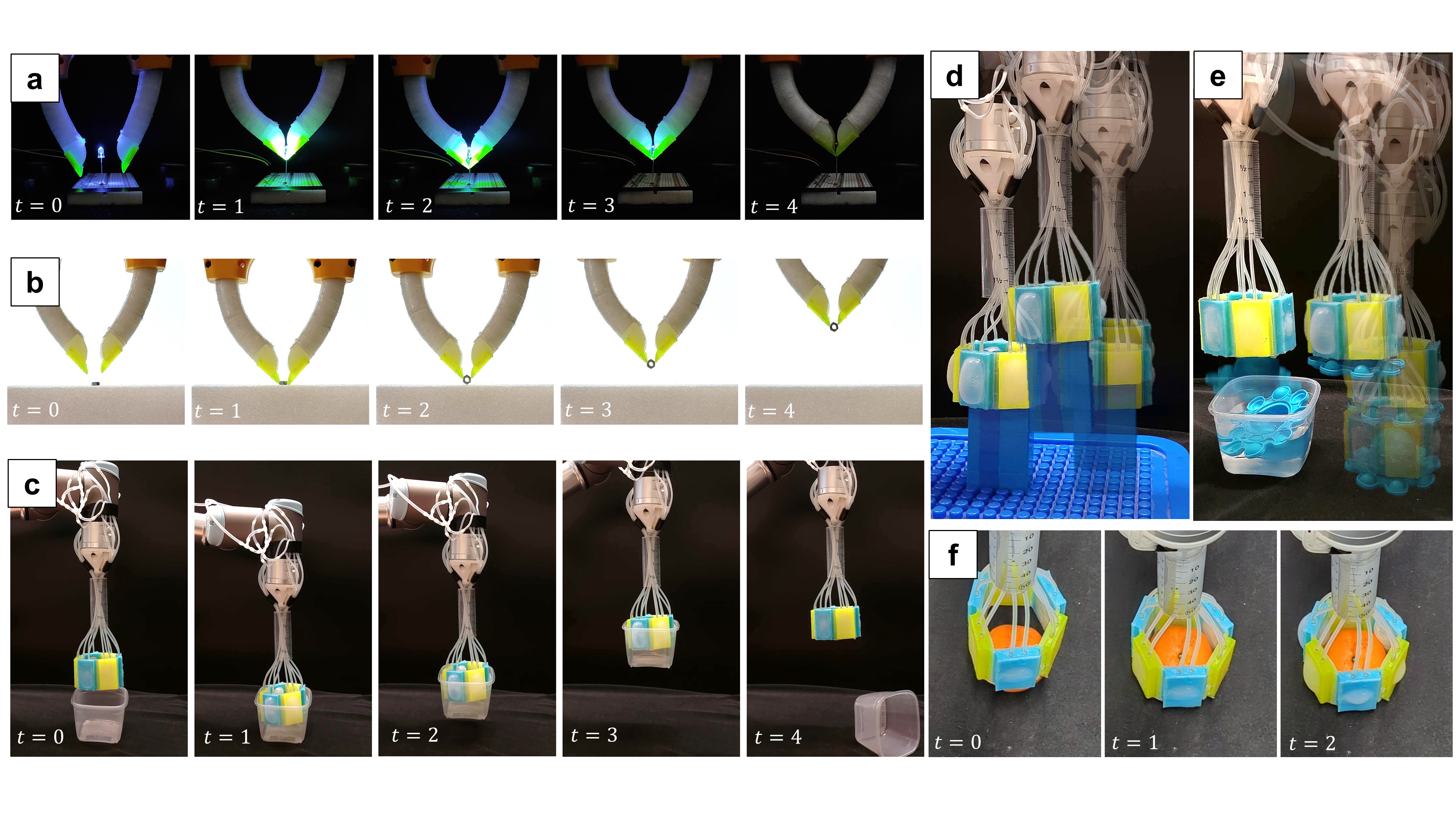}
\caption{Hybrid grippers manufactured using our proposed technique. One is a human fingernail bioinspired hybrid gripper. Similar to humans, we created a rigid nail plate connected to an underlying porous mesh made by our proposed method, with the aim of mimicking the natural bond that occurs between the rigid nail and the soft nail bed in humans once silicone rubber penetrates the porous segment. The addition of the rigid nail to a soft gripper allows the manipulation of a wide range of objects, even small ones like an LED or an M2 nut (\textbf{a} and (\textbf{b}), respectively). The other demonstrator is a hexagonal inflatable gripper that can manipulate objects from its outer (\textbf{c}) and inner sides (\textbf{d},(\textbf{e})). Underextrusion was also used as a bonding method to connect the segments of the gripper through soft hinges, which provided additional adaptability to multiple objects of varying dimensions (\textbf{f}).}
\label{fig:stream}
\end{figure}

Two hybrid grippers were designed and made for this project using our proposed technique to secure bonding between rigid and soft materials.
Inspired by the dense connective tissues found in vertebrates, we reproduced its interwoven fibrous structure using the proposed technique to connect a soft bending actuator to a rigid nail, thus recreating the bonding between the human nail and its underlying soft skin layer. Similarly to humans, adding the nail allows the hybrid gripper to grasp a wide range of objects of small dimensions, as shown in Fig.\ref{fig:stream}a-b.

As proven by the ballooning pressure tests, our method provides a novel bonding solution when fabricating inflatable structures, as it is able to withstand higher pressures when compared to other alternatives, such as silicone glue. Therefore, this can be exploited, for instance, when designing hybrid inflatable systems such as the gripper shown in Fig.\ref{fig:stream}c-f. The gripper is characterized by two different inflatable surfaces, which can be actuated simultaneously or separately, thus allowing to pick objects from both sides, as shown in Fig.\ref{fig:stream}c-f. The gripper is not only able to pick up objects of different shapes and sizes thanks to the softness of its soft inflatable paddings, but its rigid body provides a structural integrity that allows it to lift heavier objects. Furthermore, thanks to its excellent bonding properties under tensile loads, the proposed technique was also utilized in designing the soft joints that connect the different taxels of the gripper together. The use of soft joints confers great adaptability to the hybrid gripper, thus allowing it to adjust to objects of different shapes without rupture, as shown in Fig.\ref{fig:stream}d. (Supplementary video S4)

\section{Conclusion}

Inspired by the porous structure of fibrous connective tissues, this study investigated underextrusion in FDM 3D printers as a solution to the challenge of bonding soft and rigid materials in the fabrication of hybrid robots. The results of various bonding experiments demonstrate that the underextrusion interface can withstand higher loads than the traditional adhesive methods used in soft robotics. Specifically, with 30\% underextrusion, we achieved 106\% higher lap shear force and 226\% higher peeling force compared to commercial Sil-poxy adhesives when using Ecoflex 00-10. Furthermore, the proposed method provides a better bonding solution when using more viscous rubbers such as Dragon Skin 10, as it can withstand lap shear forces and peel forces that are 68\% and 281\%  higher, respectively, than those recorded when using silicone-based adhesives as a bonding option. 
Additionally, in pressure tests critical for hybrid pneumatic actuators, our method achieved four times higher pressure and twice the expansion, resulting in improved actuator performance. Moreover, using 3D printing for bonding enables the creation of more complex structures that are difficult to achieve with adhesives.

For future work, since failures often occur at the bonding point in the soft material due to stress concentration, exploring more transitional printing techniques that mimic natural tissues with a gradient in stiffness from rigid to soft materials could be beneficial. Additionally, experimenting with other 3D printing materials, such as ABS and PETG, which are prone to oozing and may provide more anchor points due to unwanted material being deposited created more fiber points. Therefore, the combination of more fibers at the bonding points with stronger material properties of these materials compared to PLA could improve bonding. Furthermore, incorporating functional materials like conductive or magnetic filaments could enhance the bonding point, allowing for the integration of sensors and actuators, thereby adding new functionalities to hybrid robotic systems.

\section{Methods}

\subsection{Biology and Inspiration}

Dense connective tissue is one of the most important components in most living organisms, thanks to its numerous mechanical functions of support, protection, and bonding between different structures. Depending on the predominance of fibers that compose it, dense connective tissue can be differentiated as fibrous or elastic. Additionally, a further division can be made based on the arrangement of the fibers that compose the tissue. Dense regular connective tissue is characterized by aligned parallel bundles of fibers, which are consequently grouped in an organized fashion, thus conferring anisotropy, as it can be seen in Fig. 1A. This tissue type can be found in multiple biological structures, such as ligaments and tendons. \cite{kamrani2019anatomy} On the other hand, the irregular, dense connective tissue is formed by an interwoven array of fibers, and it can be found in different anatomical areas that are subjected to multi-directional mechanical stresses, such as the pericardium, which is a fibrous inextensible tissue that surrounds the heart. Other than providing mechanical support to the organs, this tissue is also important for the bonding between rigid and soft structures, such as bones and skeletal muscles, which happens in the periosteum, a membrane that covers the outer surface of all bones. \cite{aaron2012periosteal,kamrani2019anatomy}. 

For instance, the so-called Sharpey's fibers, which comprise the external fibrous layer of the periosteum, are able to connect the latter with the bone by directly penetrating the circumferential and interstitial lamellae of the bone tissue. Moreover, the same fibers connect muscle tissue to the periosteum, thus completing the link between the two tissues. \cite{aaron2012periosteal,kamrani2019anatomy}.

\subsection{Controlled Weaving of Microfibers by Under-Extrusion}

In this work, we exploit commercial FDM printing to fabricate micro fibers similar to fibrous connective tissues in nature at any desired location of printed structure. Specifically, we were inspired by a common issue in the FDM printing process known as oozing, which creates undesired microfibers around the printed component, and it is usually caused by a dirty or partially clogged nozzle traveling in free space in between printing spots. This travel can stretch the liquid contamination at the tip and create a miniature fiber much smaller than the nominal size of the nozzle. On the other hand, similar thinning behavior can be seen in another FDM printing issue called under extrusion, during which the extruder fails to supply a sufficient volume of material to the nozzle, thus resulting in the stretching of the deposited material and thinning of printed lines. 
Under-extrusion is usually seen as a problem in additive manufacturing as it can lead to gaps or missing layers, thus directly affecting the quality and integrity of the component. Understanding how these fibers are created and exploiting them in a controlled manner can solve one of the major problems found within the soft robotics community, which is related to the bonding between soft and rigid bodies. Achieving good bonding is crucial when manufacturing hybrid systems. For instance, a recurrent problem when conducting this application is related to the bonding between thermoplastics (e.g., PLA) and silicone rubber, which are generally difficult to bond due to their chemical and mechanical incompatibilities.

In this work, we purposely utilized oozing and under-extrusion to create interwoven porous structures that can act as a starting point for similarly bonding different materials to Sharpey's fibers. This can be achieved by intentionally modifying two main parameters used in additive manufacturing: flow rate percentage and print speed. In the case of commercial PLA filaments, it is generally recommended to employ print speeds between 60-80 mm/sec and a flow rate of 100-105\%, depending on the color of the filament and the model of the 3D printer.

According to the underlying theory of FDM 3D printing, it will be possible to modify the dimensions of a printed filament by purposely changing one or both of the latter two parameters. Following the law of conservation of mass, it is possible to predict the diameter of any molten polymeric filament extruded through a nozzle. When you want to extrude the material over a specific distance ($d_e$), the volume of the material flowing out of the hot nozzle must equal the volume of the raw material fed through the extruder inside of the subsequent nozzle; therefore:

\begin{equation}
    V_{in} = V_{out}
    \label{eq1}
\end{equation}

If we hypothesize that the extruded filament has a rectangular section with semicircular ends, we can then determine both variables $V_{in}$ and $V_{out}$ as:

\begin{equation}
    0.25\pi(D_f)^2  d_{in} = ((l_w-l_h)l_h +0.25\pi(l_h)^2) d_e
    \label{eq2}
\end{equation}

With $D_f$ being the diameter of the raw filament, $d_{in}$ is the length of the segment fed through the extruder to extrude over the distance $d_e$. On the other hand, $l_h$ is the chosen layer height and $l_w$ is the actual width of the extruded filament, which we can compute by inverting the previous formula:

\begin{equation}
    l_w = \frac{(0.25\pi(D_f)^2 d_{in})}{(l_h d_e)} - l_h(0.25\pi-1) [mm]
    \label{eq3}
\end{equation}

Furthermore, $d_{in}$ is typically calculated by the slicer software depending on the chosen 3D printing parameters:

\begin{equation}
    d_{in} = \frac{\gamma ((l_{wn} l_h d_e)}{(0.25\pi(D_f)^2)}  [mm]
    \label{eq4}
\end{equation}

The value $l_{wn}$ is the nominal layer width, which is commonly set equal to the diameter of the nozzle, and $l_h$ is the layer height. The factor $\gamma$ indicates the flow rate percentage that can be directly set through the slicer. By substituting the latter expression in equation \ref{eq3}, we obtain:

\begin{equation}
    l_w =   \gamma  l_{wn} - l_h(0.25\pi - 1) [mm] 
    \label{eq5}
\end{equation}

This should provide us with an approximation of the diameter of the extruded filament depending on its printing parameters, thus allowing us to predict the size of the resulting fibers obtained using the proposed method. 
 
\subsection{Manufacturing Approach}

All the samples and demonstrators in this work were manufactured using a consistent approach. First, the rigid and under-extruded portions were designed as separate parts of an assembly. Next, we modified the starting code of the printer to utilize a multi-extruder 3D toolchanger capable of printing with multiple materials. When using a multi-material FDM 3D printer, it is typically possible to set different printing parameters for each extruder. By exploiting this feature, we could adjust the printing parameters for different portions of the same component, such as reducing the flow rate percentage for the porous segment. Once the part was sliced, the resulting G-code was modified with a Python script to remove unnecessary steps created by the printer, such as extruder switches, thus reducing the overall printing time.

Once the part is 3D printed and inserted into its specific mold, silicone rubber is cast first onto the under-extruded section and subsequently placed into a vacuum chamber to facilitate the penetration of rubber within the porosity. Finally, more silicone rubber is added to the mold and subsequently cured, thus completing the manufacturing of the hybrid component.

\subsection{Experimental setup and materials}

In this work, all the samples and demonstrators were printed with a Creality Ender-5 with a Bowden extruder configuration. The samples were printed with a 0.4 mm nozzle at a 0.2 mm layer height and 210 °C. The supplementary material provides a more complete overview of the printing parameters.

First, to observe the effect of the extrusion rate on the diameter of the extruded material, we performed microscopy imaging using the Keyence VHX 7000 Digital Microscope. SEM was conducted on a JSM-7200F Field Emission Scanning Electron Microscope. We 3D printed 15 samples of 20 mm x 20 mm x 3 mm of poly-lactide acid (PLA, RS-PRO 1.75 mm) with a printing temperature of 210 $^\circ C$ with a plotting speed of 80 mm/sec. The sample featured a fully extruded bottom layer of 1 mm thickness, with the top 2 mm varying extrusion rates set at 10\%, 20\%, 30\%, 40\%, 50\%, 60\%, 80\%, and 100\%. In addition, we cast Ecoflex 00-10 (Smooth-ON) into the porous structure of half of the samples to investigate silicone penetration inside the underextrusion. We chose Ecoflex 00-10 since it is a highly soft rubber, popular in soft robotics, and difficult to bond to a rigid substrate.

Furthermore, we performed a lap shear test to evaluate the bonding efficacy between commonly used adhesives in soft robotics, such as Sil-Poxy silicone rubber adhesive (Smooth-on), and our proposed method. The lap shear adhesion test followed the ASTM D5868 standard for fiber-reinforced plastics. In this test, we aimed to analyze how three different flow rate percentages (i.e., 10\%, 30\%, and 50\%) affected the bonding between the chosen printing material (i.e., PLA) and two casting materials of different stiffnesses, Ecoflex 00-10 and Dragon-Skin 10 (Smooth-on). 

For this test, a total of 12 samples (three for each flow rate percentage and three for glue) were fabricated by deliberately underextruding a $20 mm \times 20 mm \times 2 mm$ segment of the specimen, as shown in Fig S3 of the supplementary material. A $20 mm \times 20 mm \times 2 mm$ soft strip was subsequently made by casting Ecoflex 00-10 over the printed sample with an overlapping surface equal to the one circumscribed by the resulting porous structure. The test procedure consists of pulling the free segment of the silicone rubber at 13 mm/min with a universal tensile tester (Instron 3343, Instron, USA) until rupture of the sample.

In addition, a 180-degree peeling test was conducted following the ASTM D903 standard test to compare the peeling performance for measuring the stripping strength of adhesively bonded materials, as can be seen in Fig S3 of the supplementary material. However, due to the limited print bed and the large dimensions in the standard, we scaled down the dimensions and the experiment speed by six times. The influence of the flow rate percentage variation on the bond strength when using the proposed technique was compared to the bond strength recorded using a silicone adhesive. Therefore, the same number of samples as the lap shear test was manufactured for the peel test.

Circular samples with a $50mm$ outer diameter were made for the ballooning pressure tests. A $2mm$ thick external circular crown section with an inner diameter of 40 mm was printed on top of the rigid PLA substrate following our proposed underextrusion method. Similar to the bonding tests, Ecoflex 00-10 and Dragon-skin 10 (Smooth-on) were the chosen casting material in this case.
This test evaluated the pressure withstood by hybrid inflatable structures manufactured with our method compared to inflatables made using Sil-Poxy™ as an adhesive.

All the samples were inflated with a syringe (Henke Sass Wolf 50 ml) system controlled via a 3D printer driver board (Bigtreetech
Octopus v1.1 3D printer board, 8 stepper driver outputs) and the pressures were recorded with MPX5100DP air pressure sensors (NXP Semiconductors, The Netherlands). Through the control driver, we pumped 10ml/sec. At first, the test was conducted until the resulting inflated structure leaked. This was done to determine not only the maximum pressure that the samples could withstand but also the dimensions of the obtained inflated balloons, which was obtained by analyzing the test videos. Moreover, following the initial pressure results, we chose the 30$\%$ sample with Ecoflex 00-10 due to the high deformation and high usage of this material in the soft robotics community. We did a cyclic test for 1000 cycles with 80\% of the maximum tolerable pressure recorded in the initial experiments.

\subsection{Demonstrators Design}

We developed two demonstrations to showcase the application of rigidity in soft robotics and the usefulness of strong bonding between rigid and soft materials. 

First, similar to animals that utilize claws and nails to enhance their manipulation capabilities, we added a 3D-printed nail to facilitate the soft actuators in grasping small objects such as LEDs or nuts. This nail design comprises a rigid component emulating the outer structure of the nail, known scientifically as the nail plate, and an underextruded segment mimicking the connection between the connective fibers of the nail plate which is made of Keratin (keratinous) and the underlying soft nail bed. Integration of this nail was achieved by slight modifications to the molds of a common fiber-reinforced bending actuator, enabling the insertion of the nail and facilitating one-time casting of the finger to bond with the nail. Such bonding allows for more durability and higher load capacity than merely adhering to the actuator.   

In another demonstration, we designed a ballooning gripper with a rigid 3D printed structure, aiming to harness rigid materials' load capacity while benefiting from soft materials' adaptability to manipulate various objects. This design comprises panels featuring ballooning actuators on both sides, each cast separately. Subsequently, these panels were interconnected using silicone rubber in a hexagonal configuration to create soft bending joints between the rigid panels. This hexagonal structure not only retains the rigidity and adaptability of the ballooning gripper but also can deform into various shapes to accommodate different objects. Furthermore, utilizing underextrusion instead of conventional adhesives yields higher pressure values, enhancing the gripper's grasping capabilities.

\section{Supplementary Materials}

The supplementary materials for this paper can be found in \href{https://drive.google.com/drive/folders/1lllGRtQczMeEhYPQQkVfIFT1SeUSuIUQ?usp=drive_link}{supplementary materials}

\bibliographystyle{reference}
\bibliography{reference}

\begin{thebibliography}{10}
\providecommand{\url}[1]{#1}
\csname url@samestyle\endcsname
\providecommand{\newblock}{\relax}
\providecommand{\bibinfo}[2]{#2}
\providecommand{\BIBentrySTDinterwordspacing}{\spaceskip=0pt\relax}
\providecommand{\BIBentryALTinterwordstretchfactor}{4}
\providecommand{\BIBentryALTinterwordspacing}{\spaceskip=\fontdimen2\font plus
\BIBentryALTinterwordstretchfactor\fontdimen3\font minus
  \fontdimen4\font\relax}
\providecommand{\BIBforeignlanguage}[2]{{%
\expandafter\ifx\csname l@#1\endcsname\relax
\typeout{** WARNING: IEEEtran.bst: No hyphenation pattern has been}%
\typeout{** loaded for the language `#1'. Using the pattern for}%
\typeout{** the default language instead.}%
\else
\language=\csname l@#1\endcsname
\fi
#2}}
\providecommand{\BIBdecl}{\relax}
\BIBdecl

\bibitem{nature}
\BIBentryALTinterwordspacing
M.~H. Dickinson, C.~T. Farley, R.~J. Full, M.~A.~R. Koehl, R.~Kram, and
  S.~Lehman, ``How animals move: An integrative view,'' \emph{Science}, vol.
  288, no. 5463, pp. 100--106, 2000. [Online]. Available:
  \url{https://www.science.org/doi/abs/10.1126/science.288.5463.100}
\BIBentrySTDinterwordspacing

\bibitem{KIM2013287}
\BIBentryALTinterwordspacing
S.~Kim, C.~Laschi, and B.~Trimmer, ``Soft robotics: a bioinspired evolution in
  robotics,'' \emph{Trends in Biotechnology}, vol.~31, no.~5, pp. 287--294,
  2013. [Online]. Available:
  \url{https://www.sciencedirect.com/science/article/pii/S0167779913000632}
\BIBentrySTDinterwordspacing

\bibitem{rus2015design}
D.~Rus and M.~T. Tolley, ``Design, fabrication and control of soft robots,''
  \emph{Nature}, vol. 521, no. 7553, pp. 467--475, 2015.

\bibitem{OctopusarmAli}
\BIBentryALTinterwordspacing
B.~Mazzolai, A.~Mondini, F.~Tramacere, G.~Riccomi, A.~Sadeghi, G.~Giordano,
  E.~Del~Dottore, M.~Scaccia, M.~Zampato, and S.~Carminati, ``Octopus-inspired
  soft arm with suction cups for enhanced grasping tasks in confined
  environments,'' \emph{Advanced Intelligent Systems}, vol.~1, no.~6, p.
  1900041, 2019. [Online]. Available:
  \url{https://onlinelibrary.wiley.com/doi/abs/10.1002/aisy.201900041}
\BIBentrySTDinterwordspacing

\bibitem{Octopuscecilia}
\BIBentryALTinterwordspacing
Z.~Xie, F.~Yuan, J.~Liu, L.~Tian, B.~Chen, Z.~Fu, S.~Mao, T.~Jin, Y.~Wang,
  X.~He, G.~Wang, Y.~Mo, X.~Ding, Y.~Zhang, C.~Laschi, and L.~Wen,
  ``Octopus-inspired sensorized soft arm for environmental interaction,''
  \emph{Science Robotics}, vol.~8, no.~84, p. eadh7852, 2023. [Online].
  Available: \url{https://www.science.org/doi/abs/10.1126/scirobotics.adh7852}
\BIBentrySTDinterwordspacing

\bibitem{gecko}
P.~Glick, S.~A. Suresh, D.~Ruffatto, M.~Cutkosky, M.~T. Tolley, and A.~Parness,
  ``A soft robotic gripper with gecko-inspired adhesive,'' \emph{IEEE Robotics
  and Automation Letters}, vol.~3, no.~2, pp. 903--910, 2018.

\bibitem{das2023earthworm}
R.~Das, S.~P.~M. Babu, F.~Visentin, S.~Palagi, and B.~Mazzolai, ``An
  earthworm-like modular soft robot for locomotion in multi-terrain
  environments,'' \emph{Scientific Reports}, vol.~13, no.~1, p. 1571, 2023.

\bibitem{Snakeskin}
\BIBentryALTinterwordspacing
A.~Rafsanjani, Y.~Zhang, B.~Liu, S.~M. Rubinstein, and K.~Bertoldi, ``Kirigami
  skins make a simple soft actuator crawl,'' \emph{Science Robotics}, vol.~3,
  no.~15, p. eaar7555, 2018. [Online]. Available:
  \url{https://www.science.org/doi/abs/10.1126/scirobotics.aar7555}
\BIBentrySTDinterwordspacing

\bibitem{fleming2015building}
A.~Fleming, M.~G. Kishida, C.~B. Kimmel, and R.~J. Keynes, ``Building the
  backbone: the development and evolution of vertebral patterning,''
  \emph{Development}, vol. 142, no.~10, pp. 1733--1744, 2015.

\bibitem{hamrick2001development}
M.~W. Hamrick, ``Development and evolution of the mammalian limb: adaptive
  diversification of nails, hooves, and claws,'' \emph{Evolution \&
  Development}, vol.~3, no.~5, pp. 355--363, 2001.

\bibitem{culha2017design}
U.~Culha, J.~Hughes, A.~Rosendo, F.~Giardina, and F.~Iida, ``Design principles
  for soft-rigid hybrid manipulators,'' in \emph{Soft Robotics: Trends,
  Applications and Challenges: Proceedings of the Soft Robotics Week, April
  25-30, 2016, Livorno, Italy}.\hskip 1em plus 0.5em minus 0.4em\relax
  Springer, 2017, pp. 87--94.

\bibitem{buchner2023vision}
T.~J. Buchner, S.~Rogler, S.~Weirich, Y.~Armati, B.~G. Cangan, J.~Ramos, S.~T.
  Twiddy, D.~M. Marini, A.~Weber, D.~Chen \emph{et~al.}, ``Vision-controlled
  jetting for composite systems and robots,'' \emph{Nature}, vol. 623, no.
  7987, pp. 522--530, 2023.

\bibitem{Hybrid_stokes}
\BIBentryALTinterwordspacing
A.~A. Stokes, R.~F. Shepherd, S.~A. Morin, F.~Ilievski, and G.~M. Whitesides,
  ``A hybrid combining hard and soft robots,'' \emph{Soft Robotics}, vol.~1,
  no.~1, pp. 70--74, 2014. [Online]. Available:
  \url{https://doi.org/10.1089/soro.2013.0002}
\BIBentrySTDinterwordspacing

\bibitem{nail_soft}
\BIBentryALTinterwordspacing
S.~Jain, S.~Dontu, J.~E.~M. Teoh, and P.~V.~Y. Alvarado, ``A multimodal,
  reconfigurable workspace soft gripper for advanced grasping tasks,''
  \emph{Soft Robotics}, vol.~10, no.~3, pp. 527--544, 2023, pMID: 36346280.
  [Online]. Available: \url{https://doi.org/10.1089/soro.2021.0225}
\BIBentrySTDinterwordspacing

\bibitem{crossiant}
\BIBentryALTinterwordspacing
L.~Gerez, C.-M. Chang, and M.~Liarokapis, ``Employing pneumatic, telescopic
  actuators for the development of soft and hybrid robotic grippers,''
  \emph{Frontiers in Robotics and AI}, vol.~7, 2020. [Online]. Available:
  \url{https://www.frontiersin.org/articles/10.3389/frobt.2020.601274}
\BIBentrySTDinterwordspacing

\bibitem{Combustion}
\BIBentryALTinterwordspacing
N.~W. Bartlett, M.~T. Tolley, J.~T.~B. Overvelde, J.~C. Weaver, B.~Mosadegh,
  K.~Bertoldi, G.~M. Whitesides, and R.~J. Wood, ``A 3d-printed, functionally
  graded soft robot powered by combustion,'' \emph{Science}, vol. 349, no.
  6244, pp. 161--165, 2015. [Online]. Available:
  \url{https://www.science.org/doi/abs/10.1126/science.aab0129}
\BIBentrySTDinterwordspacing

\bibitem{saldivar2023bioinspired}
M.~Sald{\'\i}var, E.~Tay, A.~Isaakidou, V.~Moosabeiki, L.~Fratila-Apachitei,
  E.~Doubrovski, M.~J. Mirzaali, and A.~Zadpoor, ``Bioinspired rational design
  of bi-material 3d printed soft-hard interfaces,'' \emph{Nature
  Communications}, vol.~14, no.~1, p. 7919, 2023.

\bibitem{stress_conc_1}
R.~Balokhonov and V.~Romanova, ``On the problem of strain localization and
  fracture site prediction in materials with irregular geometry of
  interfaces,'' \emph{Facta Universitatis, Series: Mechanical Engineering},
  vol.~17, no.~2, pp. 169--180, 2019.

\bibitem{stress_conc_2}
M.~N. Saleh, M.~Saeedifar, D.~Zarouchas, and S.~T. De~Freitas, ``Stress
  analysis of double-lap bi-material joints bonded with thick adhesive,''
  \emph{International Journal of Adhesion and Adhesives}, vol.~97, p. 102480,
  2020.

\bibitem{stress_conc_3}
R.~L. Fernandes, S.~T. de~Freitas, M.~K. Budzik, J.~A. Poulis, and
  R.~Benedictus, ``Role of adherend material on the fracture of bi-material
  composite bonded joints,'' \emph{Composite Structures}, vol. 252, p. 112643,
  2020.

\bibitem{zhang2020modular}
C.~Zhang, P.~Zhu, Y.~Lin, Z.~Jiao, and J.~Zou, ``Modular soft robotics: Modular
  units, connection mechanisms, and applications,'' \emph{Advanced Intelligent
  Systems}, vol.~2, no.~6, p. 1900166, 2020.

\bibitem{ozioko2021sensact}
O.~Ozioko, P.~Karipoth, P.~Escobedo, M.~Ntagios, A.~Pullanchiyodan, and
  R.~Dahiya, ``Sensact: The soft and squishy tactile sensor with integrated
  flexible actuator,'' \emph{Advanced Intelligent Systems}, vol.~3, no.~3, p.
  1900145, 2021.

\bibitem{tatari2020bending}
M.~Tatari, S.~Kamrava, R.~Ghosh, H.~Nayeb-Hashemi, and A.~Vaziri, ``Bending
  behavior of biomimetic scale covered beam with tunable stiffness scales,''
  \emph{Scientific reports}, vol.~10, no.~1, p. 17083, 2020.

\bibitem{connective_tissue}
T.~Ushiki, ``Collagen fibers, reticular fibers and elastic fibers. a
  comprehensive understanding from a morphological viewpoint,'' \emph{Archives
  of histology and cytology}, vol.~65, no.~2, pp. 109--126, 2002.

\bibitem{interwoven}
B.~Wang, Y.~Hua, B.~L. Brazile, B.~Yang, and I.~A. Sigal, ``Collagen fiber
  interweaving is central to sclera stiffness,'' \emph{Acta biomaterialia},
  vol. 113, pp. 429--437, 2020.

\bibitem{kamrani2019anatomy}
P.~Kamrani, G.~Marston, and A.~Jan, \emph{Anatomy, connective tissue}.\hskip
  1em plus 0.5em minus 0.4em\relax StatPearls Publishing, 2023.

\bibitem{aaron2012periosteal}
J.~E. Aaron, ``Periosteal sharpey’s fibers: a novel bone matrix regulatory
  system?'' \emph{Frontiers in endocrinology}, vol.~3, p. 26839, 2012.

\bibitem{al2017mechanical}
O.~Al-Ketan, R.~A. Al-Rub, and R.~Rowshan, ``Mechanical properties of a new
  type of architected interpenetrating phase composite materials,'' \emph{Adv.
  Mater. Technol}, vol.~2, no.~2, p. 1600235, 2017.

\bibitem{ma2015hybrid}
R.~R. Ma, J.~T. Belter, and A.~M. Dollar, ``Hybrid deposition manufacturing:
  design strategies for multimaterial mechanisms via three-dimensional printing
  and material deposition,'' \emph{Journal of Mechanisms and Robotics}, vol.~7,
  no.~2, p. 021002, 2015.

\bibitem{rossing2020bonding}
L.~Rossing, R.~B. Scharff, B.~Ch{\"o}mpff, C.~C. Wang, and E.~L. Doubrovski,
  ``Bonding between silicones and thermoplastics using 3d printed mechanical
  interlocking,'' \emph{Materials \& Design}, vol. 186, p. 108254, 2020.

\bibitem{Nail_sem}
\BIBentryALTinterwordspacing
L.~Farran, A.~R. Ennos, and S.~J. Eichhorn, ``{The effect of humidity on the
  fracture properties of human fingernails},'' \emph{Journal of Experimental
  Biology}, vol. 211, no.~23, pp. 3677--3681, 12 2008. [Online]. Available:
  \url{https://doi.org/10.1242/jeb.023218}
\BIBentrySTDinterwordspacing

\bibitem{DOUPLIK201347}
\BIBentryALTinterwordspacing
A.~Douplik, G.~Saiko, I.~Schelkanova, and V.~Tuchin, ``3 - the response of
  tissue to laser light,'' in \emph{Lasers for Medical Applications}, ser.
  Woodhead Publishing Series in Electronic and Optical Materials,
  H.~Jelínková, Ed.\hskip 1em plus 0.5em minus 0.4em\relax Woodhead
  Publishing, 2013, pp. 47--109. [Online]. Available:
  \url{https://www.sciencedirect.com/science/article/pii/B9780857092373500035}
\BIBentrySTDinterwordspacing

\bibitem{Plateu-Rayleigh}
S.~Tamim and J.~Bostwick, ``Plateau–rayleigh instability in a soft
  viscoelastic material,'' \emph{Soft Matter}, vol.~17, 04 2021.

\end{thebibliography}

\end{document}